\documentclass{bmvc2k}
\title{C3VDReg: A Benchmark for Local-to-Local Colonoscopic Registration toward Anatomical Localization}

\addauthor{Linzhe Jiang}{linzhe.jiang.23@ucl.ac.uk}{1}
\addauthor{Jiayuan Huang}{}{1,3}
\addauthor{Sophia Bano}{}{1}
\addauthor{Matthew J. Clarkson}{}{1}
\addauthor{Zhehua Mao}{}{1}
\addauthor{Mobarak I. Hoque}{mobarak.hoque@manchester.ac.uk}{1,2}

\addinstitution{
 UCL Hawkes Institute\\
 University College London\\
 London, UK
}
\addinstitution{
 Division of Informatics, Imaging and Data Sciences\\
 University of Manchester\\
 Manchester, UK
}
\addinstitution{
 Visual Understanding Research Group\\
 Dept of Informatics, King's College London\\
 London, UK
}

\runninghead{Jiang et al.}{C3VDReg}

\newcommand{\figpanel}[2]{\parbox[t]{#1}{\centering\normalsize #2}}
\newcommand{\figpanelleft}[2]{\parbox[t]{#1}{\raggedright\normalsize #2}}
\newcommand{\best}[1]{\textbf{#1}}
\newcommand{\second}[1]{\underline{#1}}
\graphicspath{{images/}}

\begin{document}

\maketitle
\footnotetext[1]{$^*$Corresponding author: Mobarak I. Hoque (\url{mobarak.hoque@manchester.ac.uk}).}

\begin{abstract}
Anatomy-aware colonoscopic navigation requires localizing each partial endoscopic observation on a stable 3D reference. Such localization could support coverage assessment, revisited-region awareness, and, in the longer term, augmented guidance by linking intraoperative video with preoperative CT-derived anatomy. Yet rigid point cloud registration in the colon differs fundamentally from standard indoor or outdoor benchmarks: the surface is locally homogeneous, haustral folds are repetitive, observations are highly partial, and reconstructed depth is noisy and incomplete. We present C3VDReg, a dataset and benchmark derived from publicly available Colonoscopy 3D Video Dataset (C3VD) camera poses, depth maps, and CT meshes. For each frame, C3VDReg generates a source point cloud by depth map reprojection and a target point cloud by raycasting the CT mesh from the matched camera pose. The benchmark contains 10,015 viewpoint-matched partial-to-partial point-cloud pairs, including 2,088 held out test pairs, and evaluates public baselines under a shared contract: 8,192 points per cloud, source-only perturbations, a fixed source-to-target pose convention, and common pose metrics. Beyond introducing a benchmark, C3VDReg enables a systematic investigation of failure modes in colonoscopic registration. We find that high geometric overlap alone is insufficient for reliable pose recovery: despite 74.3-93.1\% ground-truth overlap between source and target observations, registration recall remains low across all evaluated methods. By analyzing overlap, pose errors, and translation decompositions, we identify translation ambiguity in repetitive tubular anatomy as the dominant bottleneck. This finding challenges the common assumption that increasing overlap or improving local correspondence quality is sufficient for accurate registration and highlights the need for stronger anatomical, geometric, and contextual constraints. Code, model checkpoints, and the C3VDReg dataset are publicly available at \url{https://github.com/linzhe001/C3VDReg}.
\end{abstract}

\section{Introduction}
\label{sec:intro}

Colonoscopy is central to colorectal cancer detection and surveillance, yet the endoscopist sees only a small local patch of the mucosal surface at any instant. Linking that live view to a stable 3D anatomical reference could support anatomy-aware navigation, coverage assessment, revisited-region awareness, and lesion site contextualization in a common coordinate system~\cite{cadi2025ct,bobrow2023colonoscopy,ma2021rnnslam,yao2021motion,nayor2017endoscopic}. When CT-derived colon geometry is available \cite{cadi2025ct}, this objective can be viewed as registering intraoperative endoscopic observations to preoperative anatomy, an important step toward augmented colonoscopic guidance.

However, this registration problem is not well captured by standard point cloud benchmarks. Classical pipelines such as ICP and feature-based matching, as well as learning-based methods such as PointNetLK, DCP, RegTR, and GeoTransformer, are predominantly studied on CAD objects, indoor RGB-D scenes, or outdoor LiDAR data~\cite{pomerleau2015review,huang2021comprehensive}. The colon differs fundamentally from these settings. Its surface is locally homogeneous, haustral folds repeat over long extents, each endoscopic frame provides only a highly partial observation, and reconstructed depth is noisy and incomplete. These properties make reliable keypoint detection difficult, create feature descriptor degeneracy and matching ambiguity, and can yield severe translation uncertainty even when local residuals look plausible.

Existing colonoscopy datasets provide important assets for depth estimation, pose estimation, reconstruction, and procedure-level perception, but they do not define a fixed benchmark for rigid point cloud registration between video-derived observations and CT-derived anatomy~\cite{bobrow2023colonoscopy,lena2026realsyncol,rau2024simcol3d,biffi2024realcolon,ju2024segcol}. More broadly, medical registration evaluation has emphasized that modality assumptions, coordinate conventions, and success criteria should be made explicit rather than hidden in implementation details~\cite{maier2013optical,haskins2020deep}. A systematic evaluation of rigid registration algorithms on tubular organs with strong repetitive structure, such as the colon, therefore remains missing.

Directly benchmarking local-to-global colonoscopic localization is important, but it entangles two subproblems. A method must first identify the correct region on a long, repetitive CT surface and then estimate the rigid transformation between the observed patch and that region. If both are merged into a single benchmark, low recall becomes difficult to interpret: the error may arise from incorrect visible-region retrieval, inaccurate geometric alignment after correct retrieval, or both. C3VDReg addresses this by isolating the local registration component under a controlled and reproducible protocol. For each C3VD frame, the source point cloud is reconstructed from the released depth map, while the target point cloud is obtained by raycasting the CT mesh from the matched camera pose. The task is therefore viewpoint-matched partial-to-partial registration, or local-to-local registration, which preserves the cross-domain gap between CT-derived and video-depth-derived point clouds while removing full-surface retrieval from the main benchmark.

\begin{figure*}[!t]
\centering
\setlength{\tabcolsep}{0pt}
\begin{tabular}{@{}c@{\hspace{0.02\linewidth}}c@{}}
\begin{minipage}[t]{0.49\linewidth}
\centering
\includegraphics[width=\linewidth]{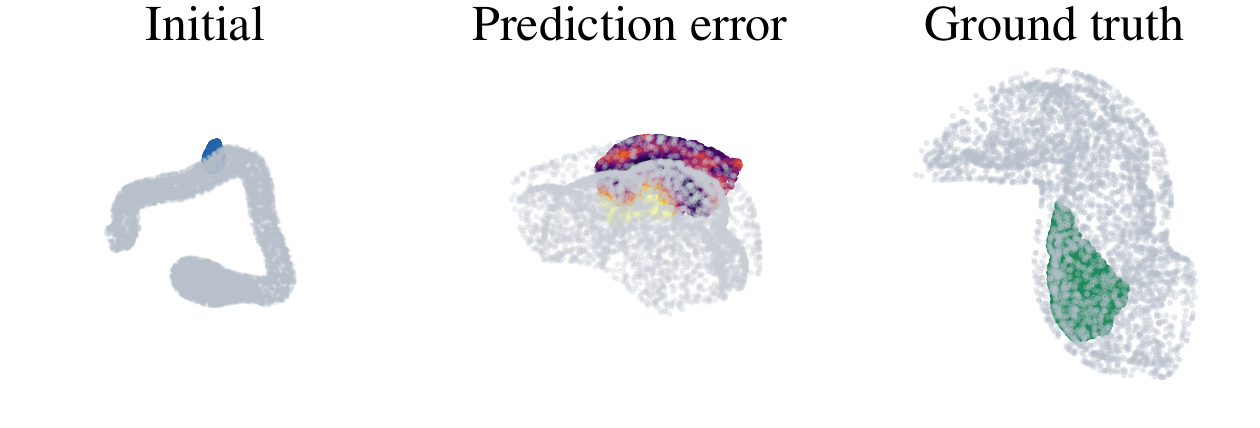}\\[-0.35em]
\figpanelleft{\linewidth}{\small(a) Sample A: local-to-global}
\end{minipage}
&
\begin{minipage}[t]{0.49\linewidth}
\centering
\includegraphics[width=\linewidth]{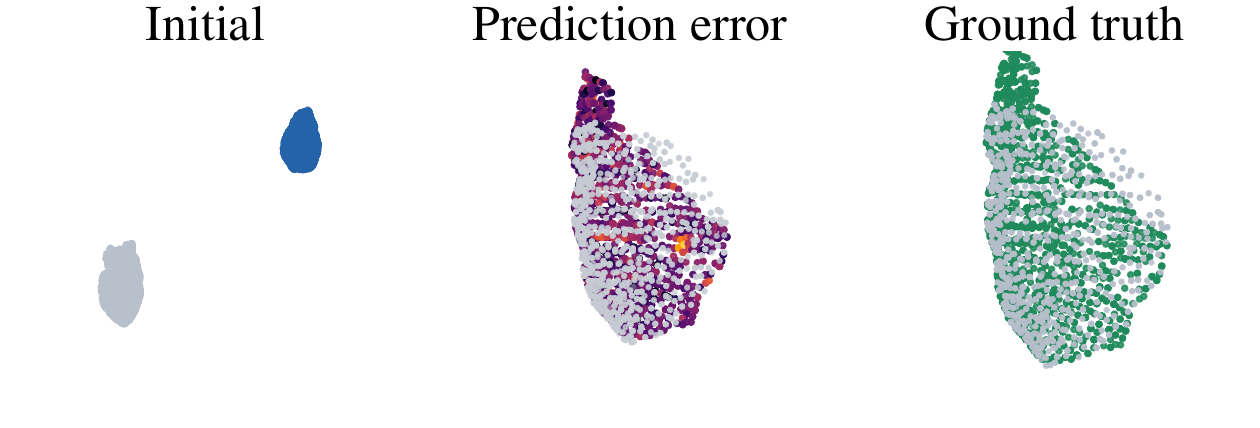}\\[-0.35em]
\figpanelleft{\linewidth}{\small(b) Sample A: local-to-local}
\end{minipage}
\\[2em]
\begin{minipage}[t]{0.49\linewidth}
\centering
\includegraphics[width=\linewidth]{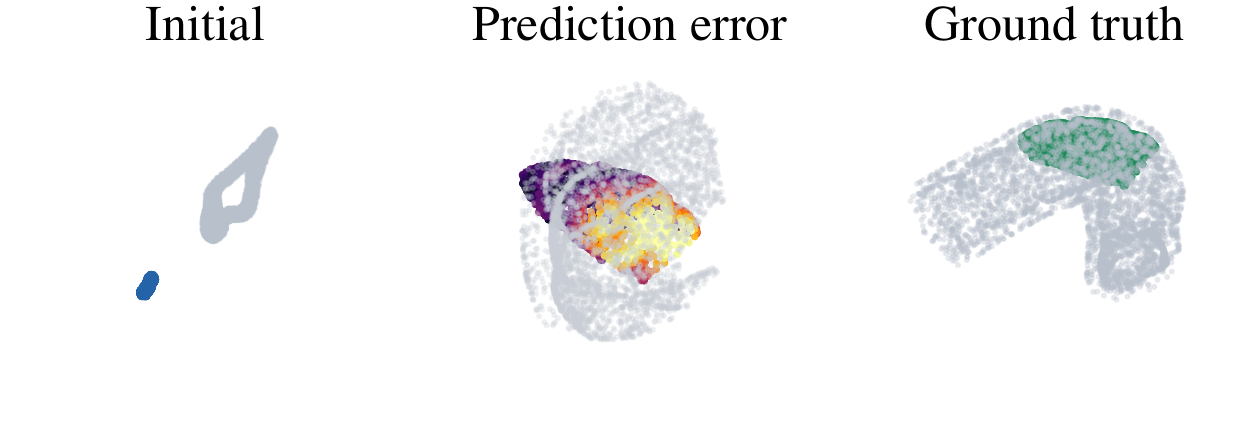}\\[-0.35em]
\figpanelleft{\linewidth}{\small(c) Sample B: local-to-global}
\end{minipage}
&
\begin{minipage}[t]{0.49\linewidth}
\centering
\includegraphics[width=\linewidth]{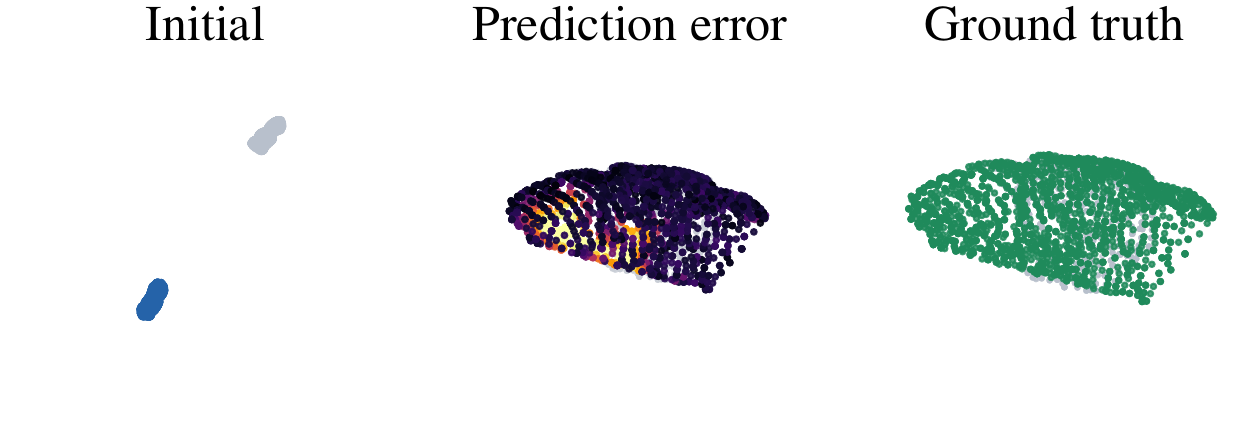}\\[-0.35em]
\figpanelleft{\linewidth}{\small(d) Sample B: local-to-local}
\end{minipage}
\end{tabular}
\caption{Local-to-global diagnostic with GeoTransformer, the strongest baseline in the main C3VDReg benchmark. The top row shows sample A and the bottom row shows sample B. The left column is local-to-global registration against the full CT surface, while the right column is local-to-local registration against the matched raycast CT surface. The poorer full-surface results motivate C3VDReg as a controlled benchmark for local registration before full-surface localization.}
\label{fig:local_to_global_scope}
\end{figure*}

Figure~\ref{fig:local_to_global_scope} illustrates why this distinction matters. We apply GeoTransformer~\cite{qin2022geometric}, the best-performing baseline in our main experiments, to the same source frames under two target definitions. With the full CT surface as target, the method must perform local-to-global matching and the examples show severe pose errors. With the matched raycast surface as target, the visible CT region is specified and the same model achieves substantially better local alignment. The figure therefore motivates C3VDReg as an intermediate benchmark: before measuring full local-to-global colonoscopic localization, it is necessary to test whether current methods can reliably align corresponding local CT and video-depth observations.

Our experiments show that even this controlled setting remains unsolved. GeoTransformer is the strongest baseline, but reaches only 17.43\% strict Registration Recall at 5deg/5mm and 35.63\% at 10deg/10mm. Ground-truth overlap is already high after alignment, yet recall remains low and nonmonotonic across overlap quantiles. The dominant bottleneck is therefore not simply missing shared surface, but translation ambiguity caused by smooth and repetitive tubular anatomy.

Our contributions are:
\begin{enumerate}
    \item We introduce \textbf{C3VDReg}, a frame-wise registration benchmark and dataset with 10,015 viewpoint-matched partial-to-partial point-cloud pairs, generated from depth map reprojection and matched-pose CT mesh raycasting.
    \item We define a fixed benchmark contract with held out test scenes, an input budget of 8,192 points, R25--90deg/T100--500mm perturbations applied only to the source, a source-to-target pose convention, and explicit pose metrics.
    \item We evaluate representative classical and learning-based rigid registration baselines and show that translation ambiguity, not low visible overlap, is the main bottleneck for frame-wise colonoscopic registration; a full-surface diagnostic further shows that direct local-to-global matching remains substantially harder.
\end{enumerate}

\section{Related Work}
\label{sec:related}

\paragraph{Colonoscopy 3D datasets and benchmarks.}
Colonoscopy datasets have broadened the scope of computer vision research in this domain, but most focus on perception tasks other than rigid registration. C3VD is particularly important because it provides colonoscopy videos together with depth maps, camera poses, surface models, and coverage annotations, making it a natural foundation for geometry-aware evaluation~\cite{bobrow2023colonoscopy}. RealSynCol and SimCol3D target depth, pose, and reconstruction using synthetic or rendered supervision~\cite{lena2026realsyncol,rau2024simcol3d}, while REAL-Colon and SegCol focus on procedure-level analysis, polyp evaluation, fold-edge segmentation, and tool-aware perception~\cite{biffi2024realcolon,ju2024segcol}. These resources are clinically relevant and increasingly realistic, but they do not define a fixed benchmark contract for frame-wise rigid registration between video-derived partial observations and CT-derived colon anatomy. In particular, source/target construction, perturbation ranges, point budgets, pose conventions, and success metrics are generally left unspecified for this task.

\paragraph{Point cloud registration.}
Rigid point cloud registration spans classical geometric optimization and modern learning-based correspondence or pose estimation. ICP remains a standard local alignment baseline~\cite{besl1992method}. Learning-based approaches include iterative feature alignment methods such as PointNetLK and PointNetLK Revisited~\cite{aoki2019pointnetlk,li2021pointnetlk}, direct pose regression or correspondence learning such as DCP~\cite{wang2019deep}, and transformer-based registration methods such as RegTR and GeoTransformer~\cite{yew2022regtr,qin2022geometric}. Surveys have highlighted the broad range of assumptions these methods make about overlap, saliency, initialization, density, and scene structure~\cite{pomerleau2015review,huang2021comprehensive}. However, their empirical evaluation is dominated by CAD objects, indoor RGB-D fragments, and outdoor LiDAR scenes. Those benchmarks do not capture the geometry of colonoscopy, where observations are highly partial, surfaces are locally smooth yet strongly repetitive, and cross-domain differences arise between CT-derived anatomy and endoscopic depth reconstruction. Strong performance on general-purpose benchmarks therefore does not necessarily translate to robust colonoscopic registration.

\paragraph{Medical registration evaluation.}
Medical registration research has repeatedly emphasized that evaluation protocols should make modality assumptions, coordinate conventions, deformation assumptions, and clinically meaningful success criteria explicit~\cite{maier2013optical,haskins2020deep}. This is especially important in medical vision, where performance can change substantially depending on how correspondences are constructed, how initialization is sampled, and whether evaluation measures global pose recovery or only local geometric fit. C3VDReg follows this principle by fixing pair construction, source-to-target pose convention, perturbation distribution, point budget, and pose-based success metrics before evaluation. In this sense, our contribution is not a new registration algorithm, but a benchmark for a previously under-evaluated setting: rigid registration on tubular colon geometry with strong repetitive structure, enabling systematic comparison of classical and learning-based methods under one reproducible protocol.

\section{Benchmark Construction and Evaluation Protocol}
\label{sec:dataset}
\label{sec:benchmark_definition}
\label{sec:protocol}

C3VDReg is a controlled intermediate benchmark for frame-wise colonoscopic registration. It has two coupled components: deterministic construction of viewpoint-matched source/target pairs from released C3VD assets, and a fixed evaluation contract shared by all baselines. This section defines both, so that performance differences reflect algorithm behavior rather than task redefinition through custom preprocessing, splits, or metrics.

\subsection{Pair Construction}
\label{sec:pair_construction}

C3VDReg is derived entirely from released C3VD videos, depth maps, camera poses, and CT surface meshes, and introduces no new human subject data collection~\cite{bobrow2023colonoscopy}. For each frame, we construct a source cloud from colonoscopic depth, a target cloud from visible CT anatomy, and the associated split and evaluation metadata.

Each registration pair links two matched but distinct measurement streams. The source point cloud $P_S$ is reconstructed from the released colonoscopic depth frame: valid pixels are unprojected with the camera intrinsics and then transformed into the global C3VD coordinate frame.

\begin{figure*}[t]
    \centering
    \includegraphics[width=\textwidth]{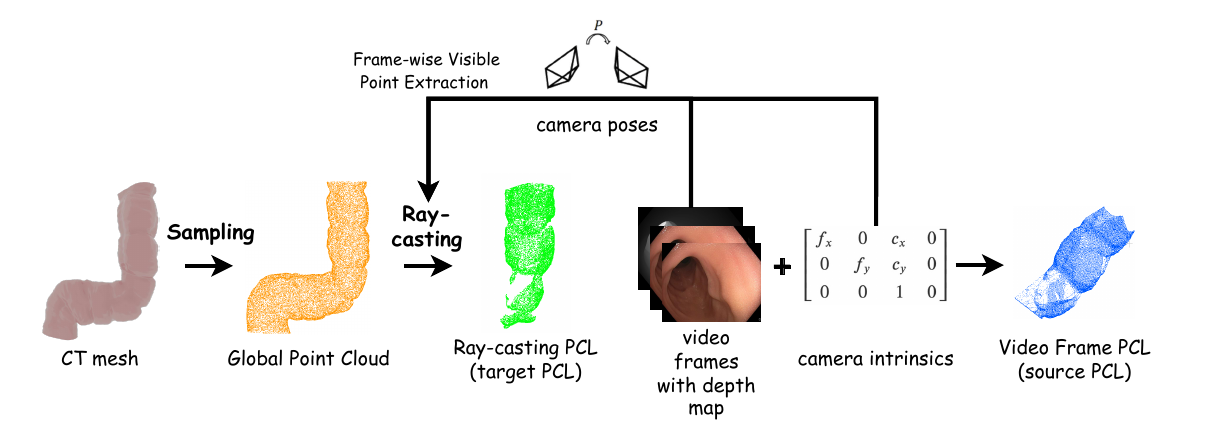}
    \caption{Frame-wise C3VDReg pair construction. For each C3VD frame, the source point cloud is generated by reprojecting the released depth map with camera intrinsics and pose. The target point cloud is generated by placing a virtual camera at the matched pose and raycasting the CT mesh to extract the visible anatomical surface. This creates viewpoint-matched partial-to-partial pairs while leaving full local-to-global retrieval outside the benchmark scope.}
    \label{fig:dataset_construction}
\end{figure*}

The target point cloud $P_T$ is generated from the CT mesh by placing a virtual camera at the same released pose, as summarized in Figure~\ref{fig:dataset_construction}. Here the CT mesh denotes the triangulated anatomical surface reconstructed from CT, and raycasting denotes intersecting each virtual camera ray with that mesh. For each image-plane sample, we retain the nearest mesh intersection along the line of sight using the Moller--Trumbore algorithm~\cite{moller1997fast}, thereby extracting the CT surface visible from the matched viewpoint. The pair is therefore viewpoint-matched but cross-domain: source and target share anatomy while differing in sensing process, density, and noise characteristics.

C3VDReg should therefore be interpreted as a benchmark for local registration after visible-region specification. The raycast target provides the oracle visible CT surface: it removes full-surface retrieval from the main task while preserving the cross-domain gap between CT-derived geometry and video-depth reconstruction. C3VDReg does not solve full local-to-global localization, viewpoint mismatch, or deformation; instead, it isolates the rigid partial-to-partial alignment problem that must be solved before those harder settings can be addressed.

\subsection{Splits and Release}
\label{sec:splits}

C3VDReg contains 10,015 generated pairs, split into 5,952 training pairs, 1,975 validation pairs, and 2,088 test pairs. To assess anatomy generalization more directly than a random pair split, the test set is defined at the scene level and contains six held-out C3VD sequences: \texttt{cecum\_t1\_a} (276), \texttt{cecum\_t3\_a} (730), \texttt{sigmoid\_t3\_b} (536), \texttt{trans\_t1\_a} (61), \texttt{trans\_t2\_b} (103), and \texttt{trans\_t4\_a} (382).

The release is intended to provide pair manifests, split files, generation scripts, evaluation scripts, and baseline configuration templates. Redistribution of derived point clouds should remain consistent with the upstream C3VD terms. We release the C3VDReg benchmark, including pair manifests, split files, generation scripts, evaluation scripts, and baseline configuration templates, at \url{https://github.com/linzhe001/C3VDReg}. The full C3VD-Raycasting-10k dataset is available at \url{https://doi.org/10.5522/04/30640043}, and the trained checkpoint bundle is available at \url{https://huggingface.co/linzher/C3VDReg-checkpoints}.

\subsection{Evaluation Protocol}

The evaluation protocol is fixed before training and testing. Each method receives a perturbed source cloud and an unperturbed target cloud, and predicts the rigid source-to-target transformation. All main results use the same point budget, perturbation distribution, pose convention, and metric suite, preventing method-specific choices from redefining the task. Table~\ref{tab:protocol} summarizes this contract.

\begin{table}[t]
\centering
\caption{Fixed C3VDReg evaluation contract used by all main experiments. The table defines the data split, point budget, perturbation range, pose convention, primary recalls, geometry diagnostics, and runtime diagnostic.}
\label{tab:protocol}
\small
\begin{tabular}{ll}
\hline
Item & Main protocol \\
\hline
Dataset source & C3VDReg \\
Task & Rigid registration between partial source and target clouds \\
Manifest & 10,015 pairs: 5,952 train, 1,975 val, 2,088 test \\
Test scenes & Six held out C3VD scenes \\
Input points & 8,192 source points and 8,192 target points \\
Perturbation & R25--90deg/T100--500mm applied only to source, no added noise \\
Primary recall & RR@5 and RR@10, defined at 5deg/5mm and 10deg/10mm \\
Error metrics & RRE in degrees, RTE in mm \\
Geometry diagnostics & Visible NN and trimmed Chamfer distance in mm \\
Runtime diagnostic & Pair latency under the shared point budget \\
\hline
\end{tabular}
\end{table}

\paragraph{Metrics.}
\label{sec:metrics}

All predictions and ground truth follow the source-to-target convention. We evaluate global pose accuracy using relative rotation error (RRE), defined as the geodesic angle between predicted and ground-truth rotations, and relative translation error (RTE), defined as Euclidean translation error in millimetres. Registration Recall at strict thresholds, abbreviated RR@5 in tables, is the fraction of pairs with RRE no larger than 5deg and RTE no larger than 5mm; RR@10 uses 10deg and 10mm. To separate orientation recovery from localization recovery, we also report marginal rotation and translation hit rates, denoted R$\leq$5 and T$\leq$5.

Mean errors summarize average performance, while medians and 90th percentiles expose typical and tail behavior. An empirical cumulative distribution function (empirical CDF) reports, for each error threshold on the $x$ axis, the fraction of evaluated pairs at or below that threshold. We do not report target registration error (TRE), because the benchmark has no independent landmark targets. Geometry diagnostics are computed after applying the predicted transform: visible nearest-neighbor (NN) distance measures the distance from a transformed source point to its closest target point, and 90\% trimmed Chamfer removes the largest residuals before computing bidirectional distance. In qualitative figures, a residual CDF uses the same cumulative view for pointwise NN residuals within one pair. These residual metrics describe local fit, but they do not replace global pose accuracy.

For prediction-independent stratification, we compute symmetric ground-truth overlap after applying the ground-truth transform. A source or target point is counted as overlapping if its nearest neighbor in the opposite cloud lies within 5mm. The final score averages the source-to-target and target-to-source fractions, and is used only to stratify the fixed test set.

Runtime columns use the same point budget for all methods. Pre. is preprocessing time, Infer is prediction time, Latency is their sum, and Lat. 90th pct. is the latency value below which 90\% of evaluated pairs finish. Eval res. and Eval alloc. are CUDA peak reserved and allocated memory during standardized batch-size-one prediction calls. Runtime is treated as a diagnostic rather than a primary success criterion, because the central benchmark question is whether a method recovers the source-to-target pose under strict RRE/RTE thresholds.

In comparable metric columns, bold marks the best value and underline marks the second best value; noncomparative, categorical, or constant columns are left unranked.

\paragraph{Baselines.}
\label{sec:baseline_initialization}

We evaluate representative methods from major registration families under the fixed C3VDReg contract: a classical local optimizer (ICP~\cite{besl1992method}), LK-style learned aligners (PointNetLK~\cite{aoki2019pointnetlk}, PointNetLK Revisited~\cite{li2021pointnetlk}, and PointNetLK-Mamba), direct pose regression or correspondence learning (DCP~\cite{wang2019deep}), and transformer-based or sparse matching methods (RegTR~\cite{yew2022regtr} and GeoTransformer~\cite{qin2022geometric}). ICP is included as a classical reference baseline evaluated on CPU rather than as a trained model. All baselines use the same split, point budget, perturbation policy, pose convention, and evaluation metrics.

\section{Benchmark Results and Analysis}
\label{sec:experiments}

We use the fixed C3VDReg protocol to answer three benchmark questions. First, how well do current registration methods perform under the main evaluation contract? Second, can their failures be explained simply by insufficient shared surface between source and target? Third, if not, what geometric failure modes remain? The analyses below show that current methods remain far from reliable, that high ground-truth overlap does not guarantee success, and that translation ambiguity in tubular anatomy is the dominant bottleneck.

\subsection{Main Results: Translation is the Bottleneck}
\label{sec:main_results}

Current public methods leave most test pairs unsolved. Under the fixed C3VDReg contract, no method reaches a reliable recall level. Aggregate recall gives the overall picture, but it is not sufficient on its own: a method may recover orientation while losing position, or succeed on a small convergent subset while failing badly elsewhere.

\begin{table*}[t]
\centering
\caption{Main leaderboard for the fixed C3VDReg protocol with R25--90deg/T100--500mm perturbations applied only to the source. RR@5 and RR@10 denote joint Registration Recall at 5deg/5mm and 10deg/10mm, respectively. Even the strongest method solves fewer than one in five test pairs at the strict threshold.}
\label{tab:main_results}
\scriptsize
\setlength{\tabcolsep}{3.2pt}
\begin{tabular}{lrrrrrrr}
\hline
Model & RR@5 & RR@10 & RRE mean & RTE mean & Visible NN & Trim CD & Latency \\
 & (\%) & (\%) & (deg) & (mm) & (mm) & (mm) & (ms) \\
\hline
GeoTransformer~\cite{qin2022geometric} & \best{17.43} & \best{35.63} & \best{10.79} & \best{44.22} & \best{1.49} & \best{2.42} & 186.8 \\
ICP~\cite{besl1992method} & \second{5.46} & 10.73 & 54.62 & 217.21 & 3.15 & 3.89 & 787.7 \\
RegTR~\cite{yew2022regtr} & 5.32 & \second{11.59} & 32.33 & 133.01 & 5.58 & 6.16 & \second{83.0} \\
PointNetLK-Mamba & 3.16 & 11.11 & \second{18.17} & \second{74.15} & 3.05 & \second{3.52} & 121.7 \\
PointNetLK Revisited~\cite{li2021pointnetlk} & 2.39 & 10.58 & 30.62 & 119.22 & \second{2.97} & 3.69 & 132.1 \\
PointNetLK~\cite{aoki2019pointnetlk} & 0.14 & 1.01 & 47.07 & 194.70 & 4.81 & 5.82 & 93.1 \\
DCP~\cite{wang2019deep} & 0.00 & 0.00 & 84.15 & 313.21 & 13.51 & 18.17 & \best{64.3} \\
\hline
\end{tabular}
\end{table*}

Table~\ref{tab:main_results} shows that the protocol is not saturated. GeoTransformer leads, but 17.43\% RR@5 and 35.63\% RR@10 leave most test pairs unsolved. ICP is a useful local reference baseline, yet its much larger mean RRE/RTE indicates that its successes are limited to a small convergent subset. DCP is the fastest method but has zero strict successes, showing that speed alone does not imply robustness under these large perturbations between partial colonoscopic clouds.

The aggregate leaderboard also obscures which part of the pose is failing. Table~\ref{tab:error_distribution} separates marginal rotation and translation hits from joint recall: GeoTransformer places 56.18\% of pairs within 5deg rotation error, but only 18.15\% within 5mm translation error. Strict RR@5 is limited by translation. ICP reaches 114/2088 strict RR@5 successes, but its median and 90th percentile RTE are 123.72mm and 570.22mm, respectively. This recall therefore reflects local convergence on a small subset rather than global robustness.

\begin{table}[!htbp]
\centering
\caption{Marginal and distributional error diagnostics for the main protocol. R$\leq$5 and T$\leq$5 are rotation only and translation only hit rates; RR@5 requires both. The large gap between GeoTransformer's rotation and translation hit rates identifies translation as the main bottleneck.}
\label{tab:error_distribution}
\scriptsize
\setlength{\tabcolsep}{3.4pt}
\begin{tabular}{lrrrrrrr}
\hline
Model & R$\leq$5 & T$\leq$5 & RR@5 & RTE med. & RTE 90th pct. & RRE med. & RRE 90th pct. \\
 & (\%) & (\%) & (\%) & (mm) & (mm) & (deg) & (deg) \\
\hline
GeoTransformer~\cite{qin2022geometric} & \best{56.18} & \best{18.15} & \best{17.43} & \best{15.86} & \best{82.12} & \best{4.29} & \best{17.16} \\
ICP~\cite{besl1992method} & 21.12 & \second{5.70} & \second{5.46} & 123.72 & 570.22 & 35.56 & 138.29 \\
RegTR~\cite{yew2022regtr} & 18.87 & 5.41 & 5.32 & 72.81 & 335.76 & 19.84 & 78.92 \\
PointNetLK-Mamba & 24.28 & 3.64 & 3.16 & 39.47 & \second{174.23} & 10.61 & \second{37.87} \\
PointNetLK Revisited~\cite{li2021pointnetlk} & \second{30.32} & 2.78 & 2.39 & \second{34.62} & 374.88 & \second{7.40} & 106.41 \\
PointNetLK~\cite{aoki2019pointnetlk} & 2.73 & 0.19 & 0.14 & 154.89 & 413.49 & 43.02 & 87.17 \\
DCP~\cite{wang2019deep} & 0.00 & 0.00 & 0.00 & 279.08 & 565.08 & 72.95 & 155.14 \\
\hline
\end{tabular}
\end{table}

\begin{figure*}[!t]
\centering
\setlength{\tabcolsep}{2pt}
\begin{tabular}{cc}
\includegraphics[width=0.48\textwidth]{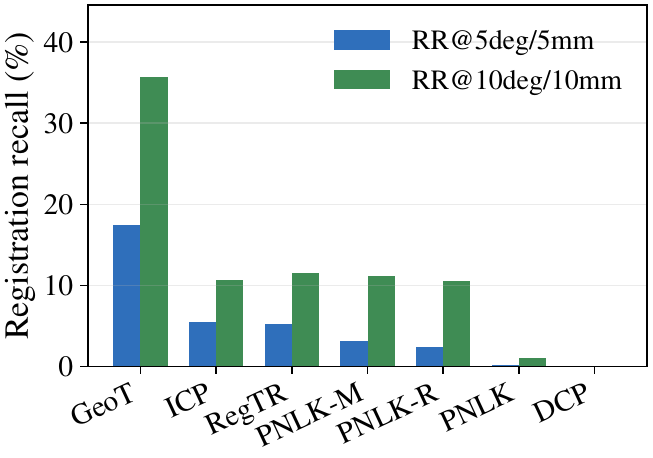} &
\includegraphics[width=0.48\textwidth]{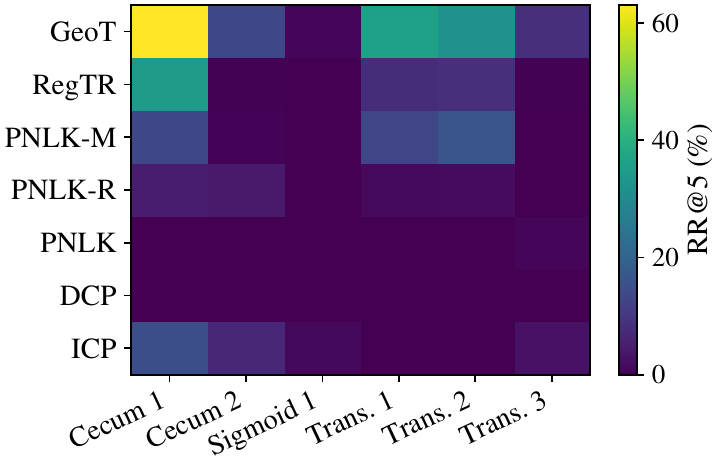} \\
\figpanel{0.48\textwidth}{\small(a) Strict registration recall} &
\figpanel{0.48\textwidth}{\small(b) Scene RR@5deg/5mm} \\[0.35em]
\includegraphics[width=0.48\textwidth]{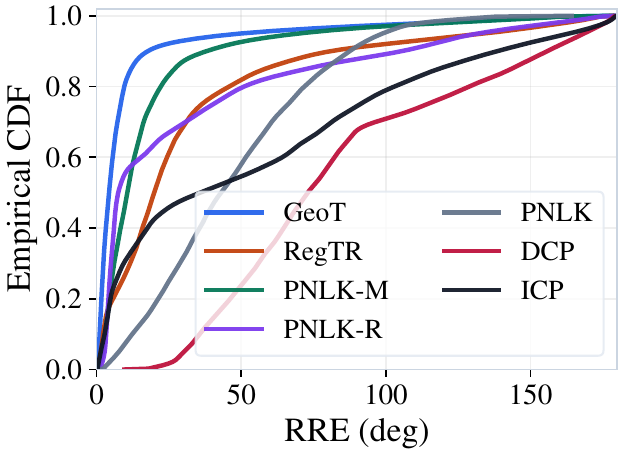} &
\includegraphics[width=0.48\textwidth]{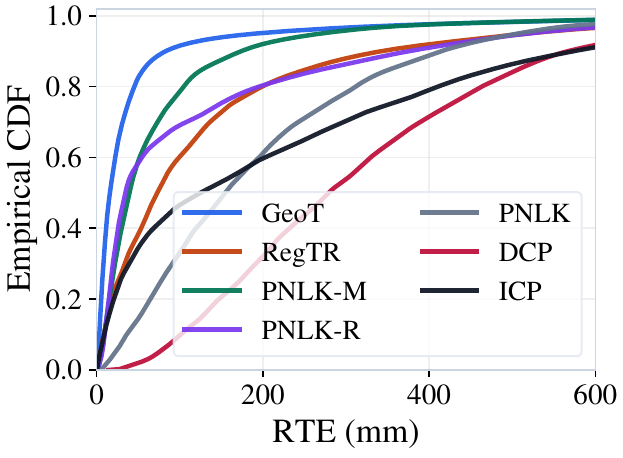} \\
\figpanel{0.48\textwidth}{\small(c) RRE empirical CDF} &
\figpanel{0.48\textwidth}{\small(d) RTE empirical CDF}
\end{tabular}
\caption{Main benchmark views. (a) Joint recall at 5deg/5mm and 10deg/10mm thresholds. (b) Scene RR@5 across held out scenes shows strong anatomy dependence. (c,d) Empirical CDFs of RRE and RTE; at any error threshold, the curve height is the fraction of test pairs below that error. Translation errors have substantially longer tails than rotation errors.}
\label{fig:main_curves}
\end{figure*}

Figure~\ref{fig:main_curves} shows the same trend at finer granularity. Scene recall is concentrated in particular held out anatomy rather than distributed evenly. GeoTransformer reaches 63.04\% RR@5 in \texttt{cecum\_t1\_a}, but only 0.93\% in \texttt{sigmoid\_t3\_b}. For compact axis labels, Figure~\ref{fig:main_curves}(b) groups the held out scenes as Cecum 1/2, Sigmoid 1, and Transverse 1/2/3. The empirical CDFs explain the leaderboard: RRE is often moderate for the strongest learned methods, while RTE retains a long tail. In this benchmark, the main difficulty is localization rather than orientation.

\subsection{High Overlap Does Not Eliminate Failure}
\label{sec:error_analysis}

A natural explanation for large translation errors is that source and target might share too little visible surface. We test this directly with a prediction-independent ground-truth overlap score. For each fixed test pair, we join per-pair results by \texttt{sample\_id} and compute symmetric ground-truth overlap after applying the ground-truth transform, using a 5mm radius and 2048 deterministic points per cloud.

\begin{table*}[t]
\centering
\begin{minipage}[t]{0.49\textwidth}
\centering
\caption{GT-overlap quantiles. RR@5 remains nonmonotonic despite high overlap; GeoT RTE is in mm.}
\label{tab:overlap_quantiles}
\scriptsize
\setlength{\tabcolsep}{1.5pt}
\begin{tabular}{@{}lrrrrr@{}}
\hline
GT overlap & \shortstack{GeoT\\[-0.15em]\cite{qin2022geometric}} & \shortstack{RegTR\\[-0.15em]\cite{yew2022regtr}} & \shortstack{ICP\\[-0.15em]\cite{besl1992method}} & PNLK-M & GeoT RTE \\
quantile & RR@5 & RR@5 & RR@5 & RR@5 & 90th pct. \\
\hline
Q1 74.3--78.1\% & \second{12.45} & 0.19 & \second{8.05} & 0.57 & 141.68 \\
Q2 78.1--83.8\% & 12.07 & \second{1.72} & 4.21 & \second{2.30} & 123.48 \\
Q3 83.8--87.8\% & \best{37.16} & \best{17.82} & \best{8.62} & \best{8.43} & \best{52.16} \\
Q4 87.9--93.1\% & 8.05 & 1.53 & 0.96 & 1.34 & \second{65.81} \\
\hline
\end{tabular}
\end{minipage}
\hfill
\begin{minipage}[t]{0.49\textwidth}
\centering
\caption{Axial/radial translation tails. Radial 90th percentile error exceeds axial error for every learned method.}
\label{tab:axial_radial}
\scriptsize
\setlength{\tabcolsep}{2.8pt}
\begin{tabular}{lrrrr}
\hline
Model & RTE & Axial & Radial & Axial \\
 & 90th pct. & 90th pct. & 90th pct. & frac. \\
 & (mm) & (mm) & (mm) &  \\
\hline
GeoT~\cite{qin2022geometric} & \best{82.25} & \best{42.62} & \best{64.77} & 0.50 \\
PNLK-M & \second{174.24} & \second{75.33} & \second{146.41} & 0.45 \\
RegTR~\cite{yew2022regtr} & 336.45 & 177.36 & 268.62 & 0.50 \\
PNLK-R~\cite{li2021pointnetlk} & 374.98 & 193.87 & 278.22 & 0.52 \\
PNLK~\cite{aoki2019pointnetlk} & 413.58 & 232.38 & 337.41 & 0.50 \\
DCP~\cite{wang2019deep} & 565.35 & 345.10 & 478.99 & 0.51 \\
\hline
\end{tabular}
\end{minipage}
\end{table*}

\begin{figure*}[t]
\centering
\setlength{\tabcolsep}{2pt}
\begin{tabular}{cc}
\includegraphics[width=0.48\linewidth]{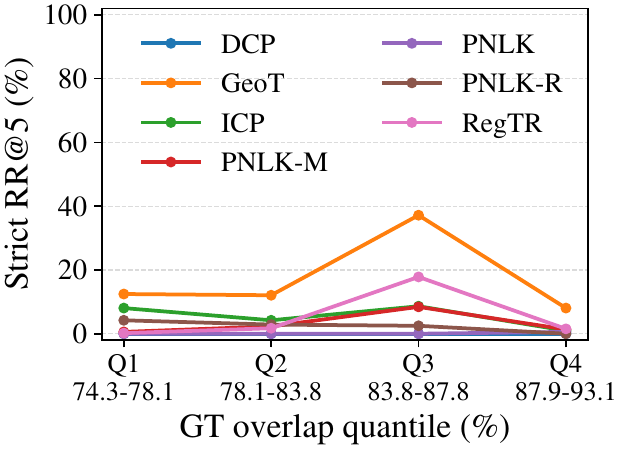} &
\includegraphics[width=0.48\linewidth]{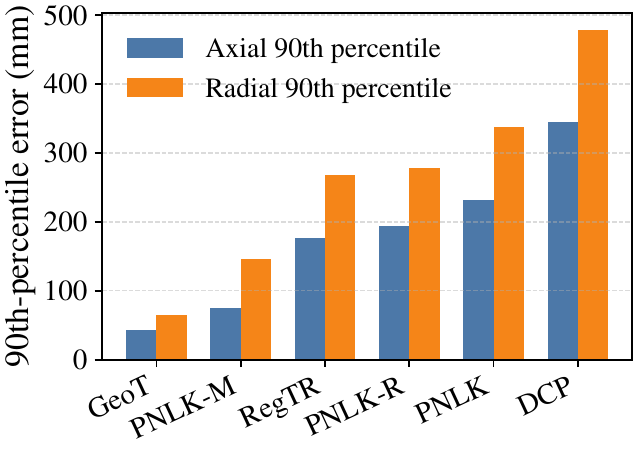} \\
\figpanel{0.48\linewidth}{\small(a) Strict RR@5 by GT overlap quantile} &
\figpanel{0.48\linewidth}{\small(b) Axial/radial translation tails}
\end{tabular}
\caption{Overlap and translation tail diagnostics. (a) Strict RR@5 is nonmonotonic across high overlap GT quantiles, ruling out missing shared surface as the dominant explanation. (b) Learned baselines retain large axial and radial translation tails, indicating both sliding along the trajectory and errors across the wall.}
\label{fig:overlap_and_tail}
\label{fig:overlap_quantile_rr}
\label{fig:axial_radial}
\end{figure*}

Table~\ref{tab:overlap_quantiles} rules out low overlap as the primary explanation in this protocol. All 2,088 test pairs have high overlap after GT alignment: the four equal size quantiles span 74.3--78.1\%, 78.1--83.8\%, 83.8--87.8\%, and 87.9--93.1\%. Yet Figure~\ref{fig:overlap_and_tail}(a) shows that strict RR@5 is nonmonotonic. GeoTransformer and RegTR improve in Q3 but fall again in Q4, while DCP remains at zero strict recall across all quantiles. High visible surface overlap does not guarantee millimetre scale global translation recovery.

\subsection{Tubular Ambiguity and the Illusion of Local Fit}
\label{sec:axial_radial}
\label{sec:qualitative}

If overlap is not the limiting variable, what drives the translation tail? A plausible anatomical failure mode is axial sliding along the colon trajectory: the model aligns a nearby tubular segment but shifts forward or backward along the lumen. We test this by decomposing each learned baseline's translation error vector into trajectory-axial and trajectory-radial components. Axial denotes motion along the local camera trajectory, while radial denotes motion across that trajectory toward or away from the colon wall. The axial direction is estimated from neighboring C3VD camera centers, so it is a trajectory-derived centerline proxy rather than a manually annotated anatomical centerline. ICP is omitted from this post hoc decomposition because the saved CPU evaluation outputs did not include per-pair transforms; its main metrics remain in Tables~\ref{tab:main_results} and~\ref{tab:error_distribution}.

The decomposition in Figure~\ref{fig:overlap_and_tail}(b) and Table~\ref{tab:axial_radial} shows that axial sliding is only part of the failure. GeoTransformer has the smallest tail, yet its axial and radial 90th percentile errors are still 42.62mm and 64.77mm. Across learned methods, the mean axial fraction stays close to 0.5, indicating that strong source perturbations produce both trajectory sliding and radial mislocalization.

\begin{figure*}[tbp]
\centering
\setlength{\tabcolsep}{0pt}
\begin{tabular}{c}
\includegraphics[width=\linewidth]{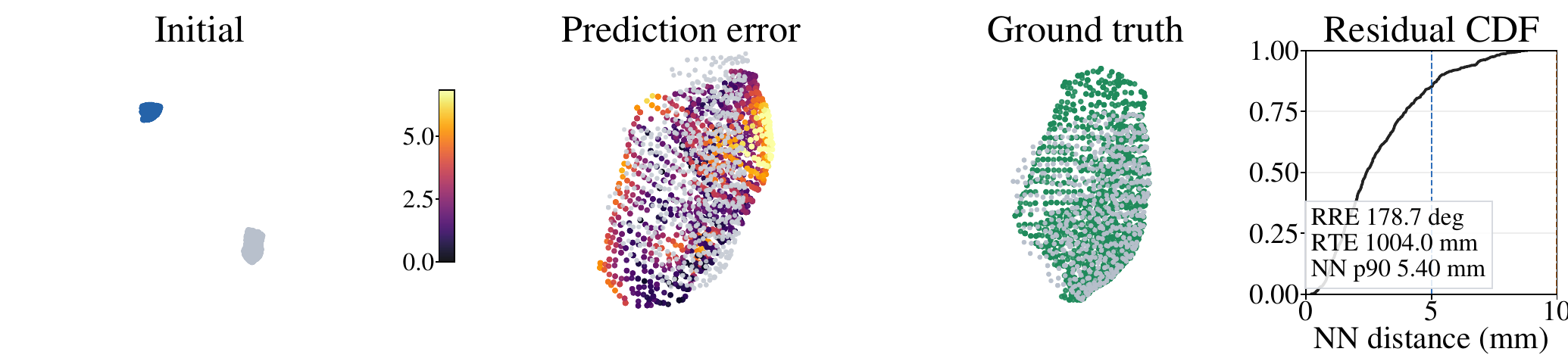}\\[-0.30em]
\figpanelleft{\linewidth}{\small(a) Low residual mismatch: NN p90 5.40 mm, RTE 1004.0 mm}\\[-0.10em]
\includegraphics[width=\linewidth]{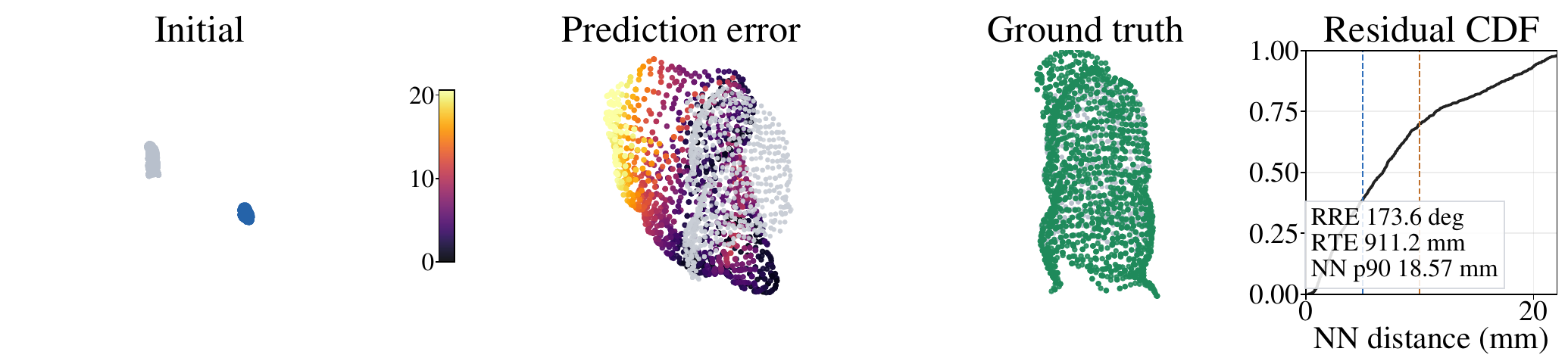}\\[-0.30em]
\figpanelleft{\linewidth}{\small(b) High residual failure: NN p90 18.57 mm, RTE 911.2 mm}
\end{tabular}
\caption{Two GeoTransformer failure regimes under C3VDReg. Columns show the initial pair, prediction error map, ground truth alignment, and residual CDF. Row (a) has low local residuals despite a globally wrong pose; row (b) has both high residuals and large pose error.}
\label{fig:qualitative}
\end{figure*}

Quantitative metrics show that local residuals and global pose accuracy can diverge, so we use GeoTransformer, the strongest baseline in C3VDReg, for a focused qualitative check. Figure~\ref{fig:qualitative} keeps two cases: a low residual mismatch, where the residual map/CDF looks locally plausible despite a wrong global pose, and a high residual failure, where local fit and pose accuracy are both poor.

The distinction is diagnostic rather than a taxonomy of all failures. The low residual case shows why residual only evaluation is insufficient: repeated tubular wall geometry can make a displaced patch look locally plausible. The high residual case is the conventional failure in which residuals and pose metrics agree. Together, the cases explain why C3VDReg reports pose metrics as the primary endpoint and uses residual distances as supporting diagnostics.

\begin{figure*}[!t]
\centering
\includegraphics[width=0.94\textwidth]{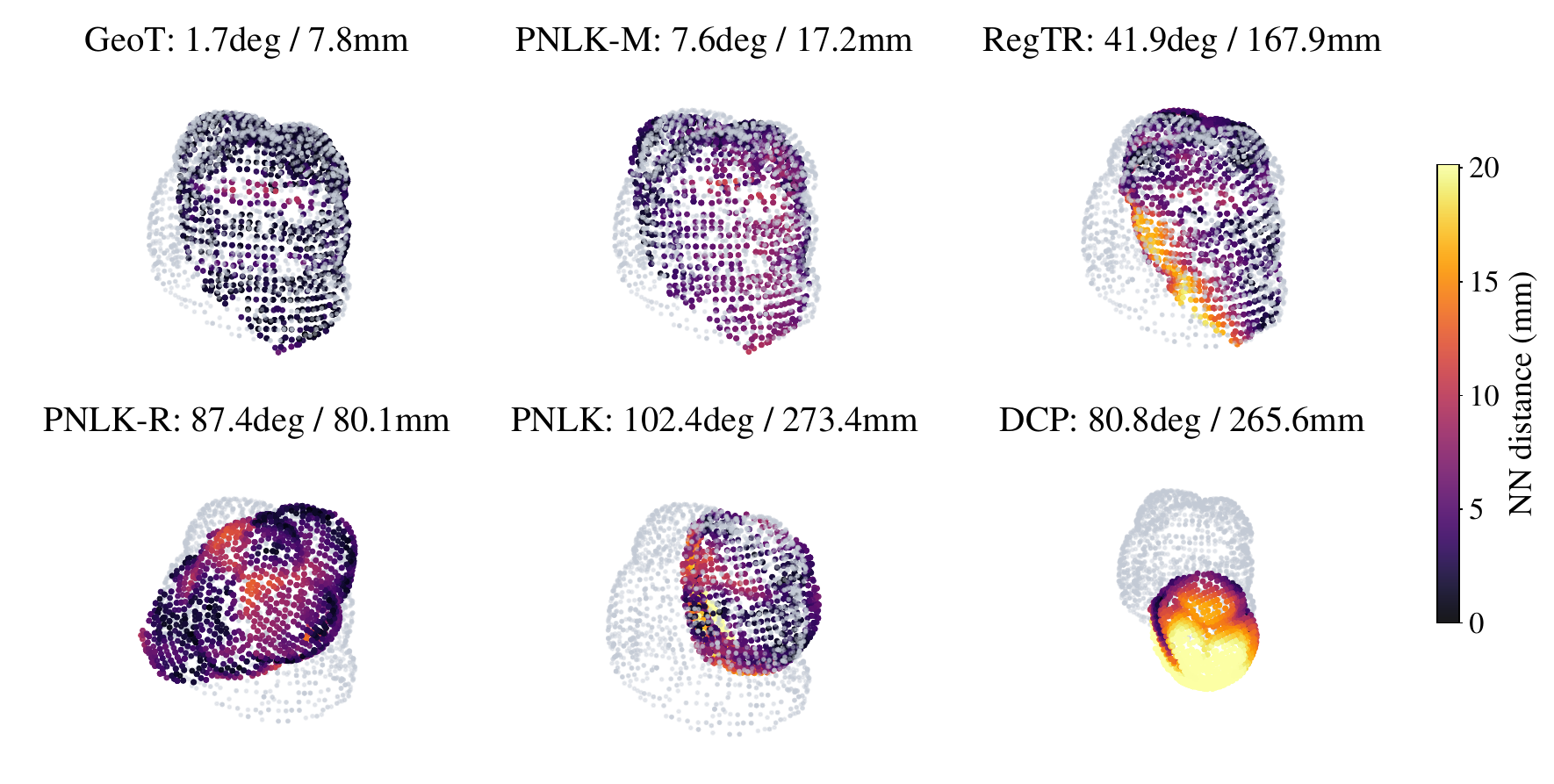}
\caption{Same sample model comparison on \texttt{cecum\_t3\_a:0133}. Target points are gray and predicted source points are colored by nearest neighbor residual with a shared scale. GeoTransformer is accurate, PNLK-M remains close, while RegTR, PNLK-R, PNLK, and DCP drift farther on the same input.}
\label{fig:same_sample_comparison}
\end{figure*}

Figure~\ref{fig:same_sample_comparison} connects the aggregate failure analysis to a single fixed test pair. Its visual ordering mirrors the translation tail ordering in Figure~\ref{fig:overlap_and_tail}(b): GeoTransformer is the most accurate, PointNetLK-Mamba (PNLK-M in the figure) is the next closest, RegTR is visibly worse, and PointNetLK Revisited, PointNetLK, and DCP drift farther on the same input. This agreement shows that the axial/radial tail metric corresponds to visible differences in global alignment quality, rather than acting only as a summary statistic.

The comparison also illustrates why model quality should not be judged by strict recall alone. RR@5 is essential because it measures whether both rotation and translation meet a deployment relevant tolerance, but it is a thresholded binary statistic: it does not distinguish a near miss from a catastrophic drift once both fail the threshold. For example, RegTR has higher RR@5 than PointNetLK-Mamba in Table~\ref{tab:main_results}, but Table~\ref{tab:axial_radial} and Figure~\ref{fig:same_sample_comparison} show that PointNetLK-Mamba has smaller translation tails and a less severe drift on the same pair. The benchmark should be read jointly through recall, RRE/RTE distributions, axial/radial tails, residual maps, and qualitative alignment, rather than using recall alone for model selection.

\subsection{The Robustness and Efficiency Tradeoff}
\label{sec:efficiency}

The preceding analyses isolate the geometric failure mode; runtime shows the practical constraint. Colonoscopic navigation is naturally online or near-online, so a useful method must be both robust to tubular ambiguity and computationally feasible. Table~\ref{tab:efficiency} shows the current gap: DCP is the fastest learned baseline at 64.3ms but has zero strict successes, RegTR is faster than GeoTransformer but much less accurate, GeoTransformer is the most robust but slower, and ICP is CPU-only and substantially slower.

\begin{table*}[!t]
\centering
\caption{Runtime and memory profile under the fixed main protocol. Timings are pairwise evaluation costs and memory columns report standardized batch size one CUDA peaks. The fastest learned baseline has zero strict recall, while the strongest baseline is slower, leaving room for methods that are both robust and near online.}
\label{tab:efficiency}
\scriptsize
\setlength{\tabcolsep}{3.5pt}
\begin{tabular}{lrrrrrrrr}
\hline
Model & Input & Train & Eval res. & Eval alloc. & Pre. & Infer & Latency & Lat. 90th pct. \\
 & pts & h/ep & MB & MB & ms & ms & ms & ms \\
\hline
GeoTransformer~\cite{qin2022geometric} & 8192 & 0.086 & \best{484.0} & \best{396.8} & \second{4.8} & 182.0 & 186.8 & 267.1 \\
ICP~\cite{besl1992method} & 8192 & -- & -- & -- & 5.0 & 782.8 & 787.7 & 946.3 \\
RegTR~\cite{yew2022regtr} & 8192 & 0.228 & \second{508.0} & 423.2 & \best{4.7} & \second{78.3} & \second{83.0} & 118.6 \\
PointNetLK-Mamba & 8192 & 0.026 & 1396.0 & 756.8 & \best{4.7} & 117.0 & 121.7 & 155.2 \\
PointNetLK Revisited~\cite{li2021pointnetlk} & 8192 & \second{0.024} & 8960.0 & 8761.6 & \second{4.8} & 127.4 & 132.1 & 162.9 \\
PointNetLK~\cite{aoki2019pointnetlk} & 8192 & \best{0.004} & 516.0 & \second{405.7} & \second{4.8} & 88.4 & 93.1 & \second{116.0} \\
DCP~\cite{wang2019deep} & 8192 & 0.324 & 6180.0 & 5341.6 & \second{4.8} & \best{59.4} & \best{64.3} & \best{67.3} \\
\hline
\end{tabular}
\end{table*}

This tradeoff matters for how the benchmark should be used. A useful method must improve recall without relying on an arbitrarily expensive matcher, and it must reduce latency without collapsing under strong perturbations. The practical target is registration that remains accurate under tubular geometric ambiguity while staying viable under colonoscopic resource constraints.

\section{Discussion, Limitations, and Future Directions}
\label{sec:discussion}

C3VDReg exposes a gap between performance on general-purpose registration benchmarks and performance on colonoscopic geometry. This benchmark is not a low-overlap stress test: all test quantiles retain high ground-truth overlap after alignment, yet strict recall remains low and nonmonotonic. The main difficulty is therefore not merely finding enough shared surface, but recovering the correct global position within smooth, repetitive tubular anatomy.

This finding helps explain why stronger local matching alone may be insufficient. The axial/radial decomposition shows that the translation tail includes both sliding along the lumen and radial mislocalization across the wall. The qualitative failure cases show the same phenomenon visually: a prediction can look locally plausible in residual space while remaining globally incorrect. Future methods may therefore need constraints beyond independent local patch matching, such as sequence context, lumen topology, anatomical priors, retrieval-aware registration, or uncertainty-aware alignment with multiple hypotheses.

C3VDReg is complementary to colonoscopy datasets focused on depth, reconstruction, detection, segmentation, or full-procedure analysis. Frame-wise localization on anatomy could support navigation, coverage assessment, revisited-region awareness, and lesion site contextualization, but C3VDReg does not itself validate clinical deployment. Its role is narrower and more technical: it isolates rigid registration between partial video-depth point clouds and visible CT target anatomy under a fixed, reproducible protocol.

The scope is intentionally controlled, and this control defines the main limitations. The raycast target provides an oracle visible CT surface, so C3VDReg does not solve full local-to-global retrieval or visible-region selection. The benchmark is rigid, viewpoint-matched, and limited to partial source and partial target clouds; it does not evaluate nonrigid deformation, unmatched viewpoints, debris, fluid, tissue motion, insufflation changes, in vivo workflow variability, or sequence-level trajectory estimation. We include ICP as a classical reference baseline, but additional robust classical methods and alternative retrieval-oriented pipelines remain to be studied. The axial/radial analysis also uses a trajectory-derived axis from C3VD camera centers rather than a manually annotated anatomical centerline. These limitations are deliberate: they keep the benchmark interpretable while identifying a clearly unsolved intermediate problem for future work.

\section{Conclusion}
\label{sec:conclusion}

We introduced C3VDReg, a frame-wise benchmark for rigid colonoscopic point cloud registration derived from released C3VD depth, pose, and CT-mesh assets. C3VDReg constructs 10,015 viewpoint-matched partial-to-partial point-cloud pairs and fixes the benchmark contract through held-out scene splits, source perturbations, a source-to-target pose convention, a shared point budget, and explicit pose-based success metrics. This design intentionally targets an intermediate problem: rigid local registration between video-depth observations and visible CT anatomy.

The benchmark results show that this intermediate problem remains unsolved. GeoTransformer is the strongest evaluated baseline, yet strict recall remains low, and the gap between rotation and translation hit rates identifies translation as the dominant bottleneck. High ground-truth overlap does not eliminate failure, and qualitative analysis shows that local residual fit can be misleading in repetitive tubular anatomy. These findings suggest that future methods will need stronger geometric and contextual constraints than local patch matching alone. C3VDReg provides a reproducible testbed for measuring progress on this step toward local-to-global colonoscopic localization and anatomy-aware colonoscopic guidance.

\clearpage
% Bibliography inlined for arXiv submission (generated with bmvc2k.bst).

\end{document}